\documentclass[10pt, a4paper]{article}

\usepackage[]{LREC/lrec2026} 

\usepackage{times}
\usepackage{boxedminipage}
\usepackage{latexsym}
\usepackage{adjustbox}
\usepackage{listings}
\usepackage{fancyvrb}
\usepackage{booktabs}
\usepackage{csquotes}
\usepackage{multirow}

\usepackage[T2A,T1]{fontenc}

\usepackage[utf8]{inputenc}
\usepackage{enumitem}

\usepackage{amsfonts}
\usepackage{amsmath}
\usepackage{multicol}
\usepackage{graphics}
\usepackage{graphicx}
\usepackage{subcaption} 

\usepackage{longtable}
\usepackage{geometry}
\usepackage{hyperref}
 \definecolor{darkblue}{rgb}{0, 0, 0.5}
  \hypersetup{colorlinks=true, citecolor=darkblue, linkcolor=darkblue, urlcolor=darkblue}

\usepackage{xstring}

\usepackage{color}

\usepackage{tabularx}

\usepackage[T1]{fontenc}

\usepackage[utf8]{inputenc}

\usepackage{microtype}

\usepackage{inconsolata}
\usepackage{multirow}
\usepackage{makecell}
\usepackage{hyperref}

\newenvironment{itemizerCompact}{\vspace{-1mm}
  \begin{itemize}
    \setlength{\itemsep}{2pt}
    \setlength{\parskip}{0pt}
    \setlength{\parsep}{0pt}
  }
{ \end{itemize}
  \vspace{-1mm}  }

\newenvironment{enumeratorCompact}{\vspace{-1mm}
  \begin{enumerate}
    \setlength{\itemsep}{2pt}
    \setlength{\parskip}{0pt}
    \setlength{\parsep}{0pt}
  }
{ \end{enumerate}
  \vspace{-1mm}  }

\newcommand{\comment}[1]{}

\newcommand{\BLU}[1]{{\color{blue} {#1}}}   

\title{Translation as Augmentation: \\Effect of Translated Data on Assessment of Difficulty }

\name{
\shortstack{
Yiheng Wu,
Jue Hou,
Roman Yangarber
}}

\address{ \comment{$^{\dagger}$Department of Digital Humanities\\
          $^{\ddagger}$Department of Computer Science \\}
          University of Helsinki, Finland\\
         \texttt{\small first.last@helsinki.fi}}

\author{Anisia Katinskaia,$^{\dagger\ddagger}$ Anh-Duc Vu,$^{\dagger\ddagger}$ Jue Hou,$^{\dagger\ddagger}$ Ulla Vanhatalo,$^{\diamondsuit}$ \\
    \textbf{Yiheng Wu,$^{\ddagger}$ Roman Yangarber$^{\ddagger}$} \\
     $^{\dagger}$Department of Computer Science, 
     $^{\ddagger}$Department of Digital Humanities \\
     $^{\diamondsuit}$Department of Finnish, Finno-Ugrian and Scandinavian studies \\
     University of Helsinki, Finland \hspace{2em} \\
     \texttt{first.last@lhelsinki.fi} \\}

\abstract{
Reliable Text Difficulty Assessment is a prerequisite for valid text simplification workflows and personalized learning applications. However, the development of robust assessment models is severely hindered by a critical bottleneck: the scarcity of expert-annotated corpora containing fine-grained difficulty levels (e.g., CEFR), particularly for lower-resource languages.  This paper addresses this data scarcity problem in the context of a low-resource European language. We propose a cross-lingual data augmentation strategy that leverages machine translation to transfer labeled resources from high-resource languages to the target low-resource language. We train BERT-based regression models to predict difficulty scores and investigate whether synthetic, translated data can effectively supplement native training sets. Our experiments demonstrate that augmenting scarce native data with machine-translated corpora significantly improves the accuracy of difficulty estimation, offering a viable solution for languages lacking extensive expert annotations.
}

\begin{document}
\maketitleabstract

\section{Introduction}

Assessment of text complexity---Difficulty Assessment---has become increasingly important for accessible communication practice, driven by policy mandates in the United States\footnote{\href{https://www.congress.gov/bill/111th-congress/house-bill/946/text/pl}{Plain Writing Act of 2010, Pub. L. 111-274}} and the European Union\footnote{\href{https://commission.europa.eu/strategy-and-policy/policies/justice-and-fundamental-rights/disability/european-accessibility-act-eaa_en}{European Accessibility Act}}.  It also plays an essential role in personalized second-language (L2) instruction \cite{nahatame2026revisiting}.  Compliance with these mandates and the personalization of L2 instruction both hinge on this shared technical bottleneck: the reliable estimation of text difficulty.  We argue that difficulty assessment is a prerequisite for the task of automatic text simplification: a simplification system cannot be meaningfully guided or evaluated without reliable metrics to determine whether its output meets the target difficulty specification.

Consequently, this paper focuses on the assessment problem.  We model this problem as a regression task, where a \textit{difficulty model} predicts a continuous score, which can then be mapped to the CEFR scale.\footnote{\href{https://www.gesetze-im-internet.de/bgg/}{Behindertengleichstellungsgesetz (BGG), Federal Republic of Germany, 2002 (amended 2016)}} Such models are critical not only for assessing learner texts but also for acting as critics in Large Language Model (LLM)-based simplification pipelines~\cite{hurst2024gpt}. A central obstacle to building robust difficulty models is data scarcity. High-quality assessment requires sizable corpora annotated by domain experts, a resource that is lacking, e.g., in many lower-resource European languages, such as Finnish. 

To address this bottleneck, we propose a cross-lingual approach: leveraging existing expert-annotated corpora from higher-resource languages and applying machine translation to generate synthetic labeled training data in the target language.
We address two {\bf Research Questions}:
\begin{enumeratorCompact}    
\item  Can training data augmented with machine-translated data from a higher-resource language improve the quality of difficulty assessment in a lower-resource language?   
\item To what extent does training on translated data improve cross-lingual generalization between the source and target languages?
\end{enumeratorCompact}

We train a BERT-based {\em regression} model to predict difficulty scores on a continuous scale. Our results indicate that using machine-translated data can substantially improve the accuracy and robustness of difficulty assessment in the target language.  The paper is organized as follows. Section \ref{sec:related work} presents an overview of related work.  Section~\ref{sec:data} describes the datasets used for training.  Section~\ref{sec:experiments} details the experimental setup and results. Section~\ref{sec:conclusion} provides conclusions and future directions.

\section{Related Work}
\label{sec:related work}
Estimating the difficulty of written text---variously referred to as readability, 
proficiency, or grade-level assessment\footnote{Throughout this paper, we use these 
terms interchangeably.}---has a long history in both educational research and NLP. 
Early formula-based approaches such as Flesch-Kincaid and the Lexile framework provide 
simple numeric difficulty scores~\cite{kincaid1975, stenner1996}, but rely on 
surface-level features and fail to capture deeper lexical or syntactic complexity. 
Subsequent supervised systems incorporated richer hand-crafted linguistic features, 
including parse depth, grammatical constructions, and word-frequency 
lists~\cite{collins-thompson-callan-2004-language, vajjala-meurers-2012-improving, 
laposhina2018-textometr}. More recent neural approaches---from hierarchical attention 
networks~\cite{azpiazu-pera-2019-multiattentive} to fine-tuned BERT 
models~\cite{martinc-etal-2021-supervised}---have substantially outperformed 
feature-based baselines, and Transformer-based models have been compared against 
feature-engineered systems for both English and Russian~\cite{sharoff2022neural}. 
Large language models have also been evaluated on readability tasks with competitive 
results~\cite{imperial-etal-2024-llm}.

Difficulty assessment plays an equally important role within text simplification 
pipelines, both as a signal for identifying complex content~\cite{gasperin2009, 
aluisio-etal-2010-readability} and as an optimization objective in rule-based 
systems~\cite{woodsend-lapata-2011-learning}. Recent work has moved toward 
feedback-driven generation, using readability classifiers with reinforcement 
learning~\cite{alkaldi-inkpen-2023-readability} or controllable generation 
conditioned on target reading level~\cite{agrawal-carpuat-2023-controlling}---a 
paradigm directly relevant to our use of a difficulty model as a critic in an 
LLM-based simplification loop~\cite{hurst2024gpt}.

A persistent bottleneck across all these approaches is the scarcity of 
expert-annotated, difficulty-labeled corpora, which is especially acute for 
lower-resource languages. Even a recent shared task on English simplification provided 
no training data~\cite{alva-manchego-etal-2025-findings}. For Finnish, we build on 
existing annotated resources~\cite{dmitrieva-konovalova-2023-creating, 
katinskaia-etal-2025-estimation} and extend them via machine translation of 
Russian-language corpora~\cite{Dmitrieva-phdthesis}, directly targeting this 
labeled-data bottleneck.

\section{Data}
\label{sec:data}

We first describe the Finnish- and Russian-language data used for training and evaluating the difficulty models.  A major challenge is the scarcity of annotated data in Finnish for prediction of difficulty.  To address this, we augment a small collection of texts in Finnish  annotated with difficulty levels---``native'' data---with a larger collection of texts in Russian that were annotated with difficulty levels, and then translated into Finnish using machine-translation (MT) models.

\subsection{Native Data}

\begin{table*}[htbp]
\centering
\begin{tabular}{l|rrrrrr|rr}
\toprule
 &  &  &  &  &  & \textbf{Native} &  & \textbf{Overall} \\
\textbf{Level} & \textbf{EL} & \textbf{TextBook} & \textbf{HS} & \textbf{YLE} & \textbf{Selko} & \textbf{total} & \textbf{MT $\leftarrow$ RU} & \textbf{total} \\
\midrule
A1   & 0   & 1   & 0   & 0   & 0   & 1   &294 & 295 \\
A1+  & 153 & 0   & 0   & 0   & 0   & 153 & 282  & 435 \\
A2   & 0   & 363 & 0   & 0   & 0   & 363& 465  & 828\\
A2+  & 0   & 0   & 0   & 0   & 0   & 0   &96 & 96 \\
B1   & 0   & 229 & 0   & 0   & 0   & 229 & 3301 & 3530\\
B1+  & 0   & 0   & 0   & 0   & 766   & 766 & 1672  & 2438\\
B2   & 0   & 163 & 0   & 0   & 0   & 163 & 834 & 997 \\
B2+  & 0   & 0   & 0   & 0   & 0   & 0   & 0    & 0    \\
C1   & 0   & 192   & 0   & 0   & 0   & 192   & 484    & 676  \\
C1+   & 0   & 0   & 715 & 703 & 0   & 1418 & 0  & 1418 \\
C2  & 0   & 175 & 0 & 0 & 0   & 175 & 29   & 204 \\
\midrule
\textbf{Total} & 153 & 1123 & 715 & 703 & 766 & 3460 & 7457 & 10917 \\
\bottomrule
\end{tabular}
\caption{Number of documents in Finnish {\em native} and machine-translated (MT) datasets by CEFR Level}
\label{tab:dataset_stats}
\end{table*}

\begin{table}[t]
\centering
\scalebox{0.78}{
\begin{tabular}{l|c|c|r|r|r}
\toprule
    \textbf{} & \textbf{} & \textbf{} & \textbf{Total} & \textbf{Average} & \textbf{Average}\\

    \textbf{Source} & \textbf{CEFR} & \textbf{Level} & \textbf{\# Docs} & \textbf{\# Words} & \textbf{\# Sent.}\\
    \midrule
    RuFoLa          & A1 & 1.0 & 301 & 136 & 8.8 \\
    Encyclop.    & A1-A2 & 1.5  & 282 & 31 & 12.3 \\
    RuFoLa          & A2 & 2.0  & 466 & 183 & 10.5 \\
    Zlatoust        & A2-B1 & 2.5  & 96 & 50 & 8.2 \\
    RuFoLa          & B1 & 3.0  & 3306 & 91 & 12.2 \\
    Zlatoust        & B1-B2 & 3.5  & 1677 & 54 & 15.8 \\
    Zlatoust        & B2 & 4.0  & 834 & 228 & 12.8 \\
    RuFoLa          & C1 & 5.0  & 485 & 363 & 14.9 \\
    RuFoLa          & C2 & 6.0  & 29 & 385 & 16.5 \\
    \bottomrule
\end{tabular}}
\caption{\label{tab:data-ru} Annotated documents in Russian.}
\end{table}

We compile a dataset for Finnish by combining various native and machine-translated sources, spanning the range of CEFR readability levels (A1--C2).  Table~\ref{tab:dataset_stats} provides an overview of the composition of this dataset.  Native data is drawn from five main sources: 
\begin{itemizerCompact}
    \item {\em Easy Language} (EL)---a collection of texts from government and NGO websites, written in ``Easy Language'' for non-native speakers; 
    \item {\em TextBook}---a collection of texts from textbooks for L2 learning, at various CEFR levels;
    \item {\em Helsingin Sanomat} (HS)---a commercial news site with the widest coverage nationally;
    \item {\em YLE}---a government news site;
    \item {\em YLE Selkouutiset} (Selko)---YLE's simplified news for non-native speakers and L2 learners.  
\end{itemizerCompact}
Each source contributes texts at different levels of linguistic complexity and genre.  {\em Easy Language} and {\em YLE selkouutiset} provide simplified Finnish materials aimed at beginners and intermediate learners, while {\em YLE} and {\em Helsingin Sanomat} offer authentic journalistic texts at advanced CEFR levels (C1–C2).  In total, the native corpus contains 4544 texts distributed across the CEFR scale.

We assign each text in this collection to one of 11 classes---these correspond to the 6 ``principal'' CEFR levels, plus 5 {\em intermediate} levels, i.e., A1+, A2+, B1+, etc. 
The rationale for introducing the intermediate levels is as follows.  
Some sources, such as textbooks, provide fine-grained assignment of the texts to the CEFR levels.  However, other sources (e.g., news sites) yield only a coarse-grained {\em estimate} of difficult.  Thus, we assume that newspaper texts are on a hypothetical level ``C'', approximately between C1 and C2.
\comment{!!!why do we use only <500 SELKO documents?  when we have thousands of them ... Answer: All table are updated }

It is also important to note that in modeling we make the assumption that the CEFR levels are {\em evenly} spaced on a linear difficulty scale.  This is done because an exact spacing of the levels on the CEFR scale is {\em latent} and not known explicitly. 
For modeling, in Section \ref{sec:model-settings}, we map these levels onto a continuous numerical scale ranging from 1 to 6, preserving their relative order while enabling regression-based prediction of text difficulty.

To address the problem of data scarcity in Finnish---especially at the lower readability levels---we augment the corpus with machine-translated (MT) texts, which have CEFR annotations in the original.  This process is described in detail in Section \ref{sec:translated-data}.
The translated subset (labeled “MT~$\leftarrow$~RU” in Table~\ref{tab:dataset_stats}) contains 8321 documents, which mirror the CEFR distribution of the native data to support balanced training. Including translated data allows us to examine whether CEFR-labeled content from a high-resource language can improve performance in a low-resource target language.

Across both native and translated sub-corpora, the dataset covers all CEFR levels (A1--C2), with a total of 12{,}865 texts. Lower levels (A1–A2+) are primarily sourced from {\em Easy Language}, {\em TextBook}, and translated materials, while mid-level texts (B1–B2) are drawn from {\em YLE selkouutiset} and corresponding MT data.  The advanced levels (C1–C2) are mainly represented by authentic Finnish news and literary texts from {\em HS} and {\em YLE}.  This distribution ensures that the dataset reflects both pedagogically simplified and naturally complex usage of Finnish, enabling robust analysis of cross-lingual transfer and translation-based augmentation across readability levels.  We compile a comprehensive Finnish dataset by combining multiple native and machine-translated sources, spanning a wide range of CEFR readability levels (A1–C2).

\subsection{Translated Data}
\label{sec:translated-data}

We use text resources in Russian that have been manually annotated for difficulty, and translate them into Finnish to augment the training dataset.  We use a collection of Russian simple-language corpora, introduced in \cite{Dmitrieva-phdthesis}.  Two corpora of annotated Russian texts, shown in Table~\ref{tab:data-ru}:\comment{!!! does not match table 2 -- the CORPORA names Answer: I added new descriptionj}
\begin{itemizerCompact}
    \item the {\em RuFoLa} corpus~\cite{laposhina2020corpus}, which contains texts from coursebooks designed for learners of Russian as a foreign language;  
    \item the {\em RuAdapt} corpus~\cite{dmitrieva-tiedemann-2021-creating}, a {\em parallel} corpus of Russian--Simple Russian, with authentic texts adapted for learners of Russian as a foreign language. In this paper, we use only the literary ({\em Zlatoust}) and encyclopedic sub-corpora  ({\em Encyclop.}). 
\end{itemizerCompact}

\begin{table}[t]
\centering
\begin{tabular}{l|rrr}
\toprule
\textbf{Split} & \textbf{FI } & \textbf{RU (MT)} & \textbf{Both} \\
\midrule                                                 
Train          & 2,332                & 5,961            & 8,293  \\
Dev            &   263                &   748            & 1,001  \\
Test           &   865                &     748            &   1,613  \\
\midrule                                                          
\textbf{Total} & 3,460                & 7,457            & 10,917 \\
\bottomrule
\end{tabular}
\caption{Data splits by source language}
\label{tab:data-split}
\end{table}

We translate the Russian texts into Finnish using models from OpusMT.\footnote{\href{https://opus.nlpl.eu/dashboard/Tatoeba-MT-models/sla-fin/opusTCv20210807\_transformer-big\_2022-09-15}{Tatoeba MT model for Slavic--Finnish.}}  It is critical to note that machine translation does not {\em guarantee} that a text in Russian will remain at the same difficulty level after translation into Finnish.  This question---under what conditions and to what extent do MT models preserve the difficulty level of the original text---deserves detailed investigation on its own; we would expect that this would depend heavily on how the MT model is trained.  However, these particular MT models with which we experiment do seem to exhibit a strong ability to preserve the difficulty level across the translation, as confirmed by manual inspection of a sample by native language experts.


\begin{table*}[t]
\centering

\begin{tabular}{lrrrrrr}
\toprule
\textbf{Sample} &
\makecell[c]{\textbf{TTR}} &
\makecell[c]{\textbf{Lexical} \\ \textbf{Density}} &
\makecell[c]{\textbf{Mean Word} \\ \textbf{Length}} &
\makecell[c]{\textbf{Mean Sent.} \\ \textbf{Length}} &
\makecell[c]{\textbf{Mean Clause} \\ \textbf{Length}} &
\makecell[c]{\textbf{POS} \\ \textbf{Diversity}} \\
\midrule
FI    & 0.6993 & 0.6930 & 7.40 & 8.65  & 19.67 & 1.90 \\
RU    & 0.8262 & 0.5934 & 5.30 & 12.71 & 10.76 & 2.04 \\
FI+MT & 0.8598 & 0.6338 & 6.61 & 9.31  & 10.68 & 1.85 \\
\bottomrule
\end{tabular}
\caption{Linguistic features for FI, RU and FI+MT samples.}
\label{tab:linguistic_features}
\end{table*}

\begin{table*}[htbp]
\centering
\begin{tabular}{clllrrrr}
\toprule
\textbf{Exp} & \textbf{Lang} & \textbf{Train \& Dev Set} &  \textbf{Test Set} & \textbf{MSE (\%)} & \textbf{RMSE (\%)} & \textbf{MAE (\%)} & \textbf{$R^2$ (\%)} \\
\midrule
1 & FI & Native  &  Native  & 5.12 & 22.63 & 12.17 & {\bf 97.30} \\
2 & FI & Native  & MT  & 146.84 & 121.18 & 102.59 & -99.71 \\
3 & FI & MT  & Native  & 155.00 & 124.50 & 109.25 & 18.24 \\
4 & FI & MT &  MT& 24.54 & 49.54 & 26.80 & 66.93 \\
5 & FI & Native + MT &  Native & 7.57 & 27.51 & 9.43 & 96.01 \\
6 & FI & Native + MT & MT   & {\bf 4.22} & {\bf 20.55} & {\bf 8.24} & 94.26 \\
\midrule  
7 & RU & Native  & Native  &  19.19 & 43.81 & 26.04 & 73.90 \\
8 & RU & Native + MT & Native & {\bf 16.72}& {\bf 40.89}&  {\bf 19.43} &  {\bf 77.26}\\
\bottomrule
\end{tabular}
\caption{Model performance on difficulty prediction for Finnish and Russian text.  MT denotes data autmentation via machine translation (Russian to Finnish). Experiments 1--8 test various training vs.~test combinations of native vs.~translated datasets.}
\label{tab:results}
\end{table*}

\comment{ 
\begin{table*}[htbp]
\centering
\resizebox{\textwidth}{!}{
\begin{tabular}{cllllrrrr}
\toprule
\textbf{Exp} & \textbf{Lang} & \textbf{Train Set} & \textbf{Dev Set} & \textbf{Test Set} & \textbf{MSE (\%)} & \textbf{RMSE (\%)} & \textbf{MAE (\%)} & \textbf{$R^2$ (\%)} \\
\midrule
4 & FI & Native  & Native  & Native  & 5.12 & 22.63 & 12.17 & {\bf 97.30} \\
6 & FI & Native  & Native  & MT  & 146.84 & 121.18 & 102.59 & -99.71 \\
5 & FI & MT  & MT & Native  & 155.00 & 124.50 & 109.25 & 18.24 \\
1 & FI & MT & MT & MT& 24.54 & 49.54 & 26.80 & 66.93 \\
3 & FI & Native + MT & Native + MT & Native & 7.57 & 27.51 & 9.43 & 96.01 \\
2 & FI & Native + MT & Native + MT & MT   & {\bf 4.22} & {\bf 20.55} & {\bf 8.24} & 94.26 \\
\midrule  
7 & RU & RU Native  & RU Native  & RU Native   & 19.19 & 43.81 & 26.04 & 73.90 \\
8 & RU & Native + MT & RU Native  & RU Native & {\bf 16.72}& {\bf 40.89}&  {\bf 19.43} &  {\bf 77.26}\\
\bottomrule
\end{tabular}}
\caption{Model performance on Finnish and Russian text difficulty prediction. MT denotes data machine-translated from Russian to Finnish. Each experiment (Exp.~1--7) tests different training and evaluation combinations of native and translated datasets.}
\label{tab:results}
\end{table*}
}

Examining the instances in the Russian corpus, we find that some instances that are too short or too long. Therefore, we removed some outlier instances, and hence the number of {\em MT ← RU} documents in Table~\ref{tab:dataset_stats} is somewhat lower than the original Russian texts in Table~\ref{tab:data-ru}.  All documents---native and translated---were split into 3 sets: training, validation, and test, as shown in Table~\ref{tab:data-split}. For the experiments with Finnish, we translated the Russian training and validation sets, and added them to the corresponding Finnish sets, to augment the limited size of the Finnish sets.  In contrast, for the experiments with Russian, we translated only the Finnish training set and incorporated it into the Russian training set, as the Russian development set was already sufficiently large.

Table \ref{tab:linguistic_features} reports six metrics of linguistic complexity, which are commonly used to characterize text difficulty:
\begin{itemizerCompact}
\item Type-Token Ratio (TTR)---is a measure of vocabulary richness,
\item Lexical Density---measures the proportion of content words, 
\item Mean Word Length---measures morphological variety,
\item Mean Sentence Length---reflects syntactic variety,
\item Mean Clause Length---reflects the structural complexity of sentences, capturing both the elaboration of clause-internal constituents and the depth of syntactic embedding,\comment{!!! proper term?, Answer I changed into another definition} 
\item POS Diversity---indicates part-of-speech variety within the text.    
\end{itemizerCompact}
  These features roughly reflect the lexical and structural properties of the texts in the dataset.  They are not used in modeling (at present), and are presented to give the reader an intuitive description of the data. \comment{!!! where is this information USED ? Answer: I want to show some statistics for this dataset. So I changed this description.  }

The main point in this Section is that the dataset augmented with translated data is 3 times larger than the original ``native'' dataset---which we hope will help train a more accurate regresion model.

\section{Experiments}
\label{sec:experiments}

To examine whether machine-translated data can improve text difficulty prediction, we train regression models on native Finnish texts and translated Russian texts. This allows us to explore the research questions: assess how MT-based data augmentation influences model performance and cross-lingual generalization in predicting document-level difficulty.

\subsection{Model settings}
\label{sec:model-settings}

We build a BERT-based regression model to predict text difficulty.  For Finnish, we use the TurkuNLP/bert-base-finnish-cased-v1 model; for Russian, we use ai-forever/ruBert-large. Both models are fine-tuned with a regression objective.

\renewcommand{\thesubfigure}{\arabic{subfigure}}
\def\picwidth{.90}
\def\pictop{23mm} 
\begin{figure*}[htbp]
    \centering
    \begin{subfigure}[b]{0.49\textwidth}
    \centering
    \includegraphics[trim=20mm 10mm 20mm \pictop, clip, width=\picwidth\linewidth]{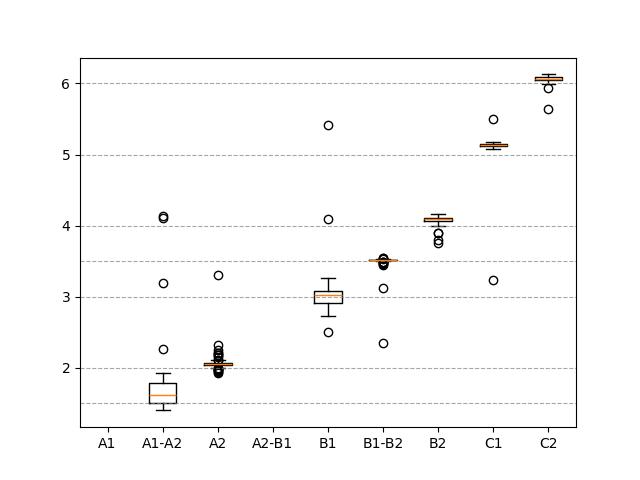}
        \caption{FI train and test on native data}
         \label{fig:boxplos-RQ1-1}
    \end{subfigure}
    \begin{subfigure}[b]{0.49\textwidth}
    \centering
    \includegraphics[trim=20mm 10mm 20mm \pictop, clip, width=\picwidth\linewidth]{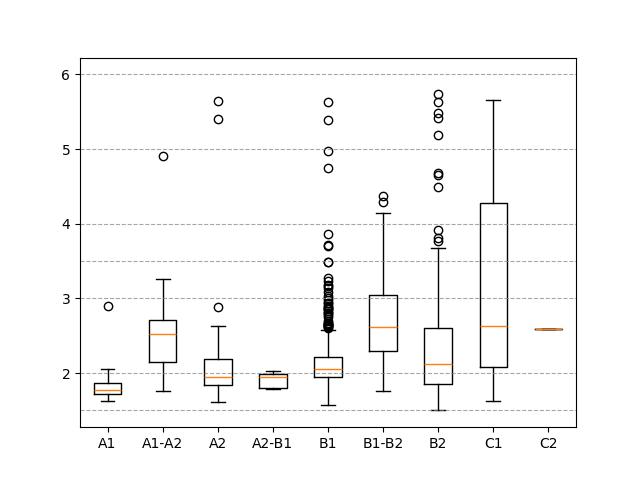}
        
        \caption{FI train on native, test on MT}
        \label{fig:boxplos-RQ1-2}
    \end{subfigure}
    \begin{subfigure}[b]{0.49\textwidth}
    \centering
    \includegraphics[trim=20mm 10mm 20mm \pictop, clip, width=\picwidth\linewidth]{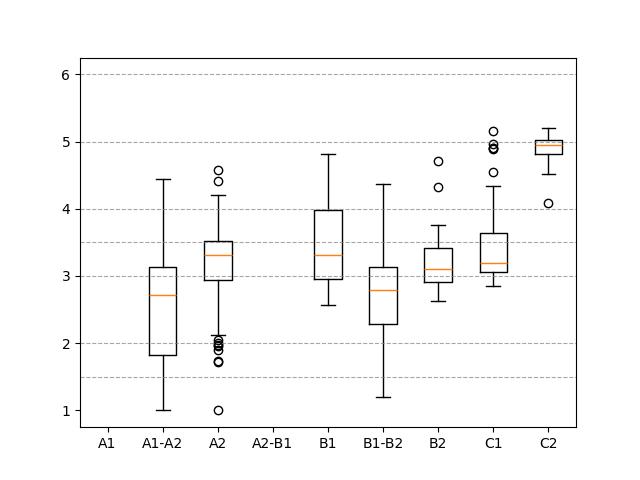}
        
        \caption{FI train on MT, test on native data}
        \label{fig:boxplos-RQ1-3}
    \end{subfigure}
    \begin{subfigure}[b]{0.49\textwidth}
    \centering
    \includegraphics[trim=20mm 10mm 20mm \pictop, clip, width=\picwidth\linewidth]{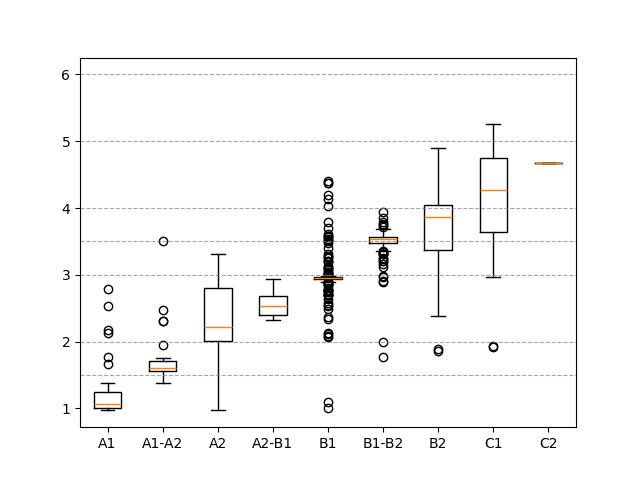}
        
        \caption{FI train and test on MT}
        \label{fig:boxplos-RQ1-4}
    \end{subfigure}
    
    \vskip\baselineskip
    \begin{subfigure}[b]{0.49\textwidth}
    \centering
    \includegraphics[trim=20mm 10mm 20mm \pictop, clip, width=\picwidth\linewidth]{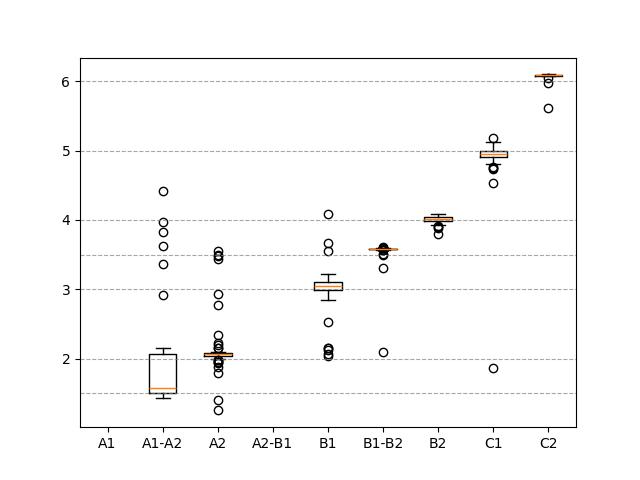}
        \caption{FI train on all data, test on native,}
        \label{fig:boxplos-RQ1-5}
    \end{subfigure}
    \begin{subfigure}[b]{0.49\textwidth}
    \centering
    \includegraphics[trim=20mm 10mm 20mm \pictop, clip, width=\picwidth\linewidth]{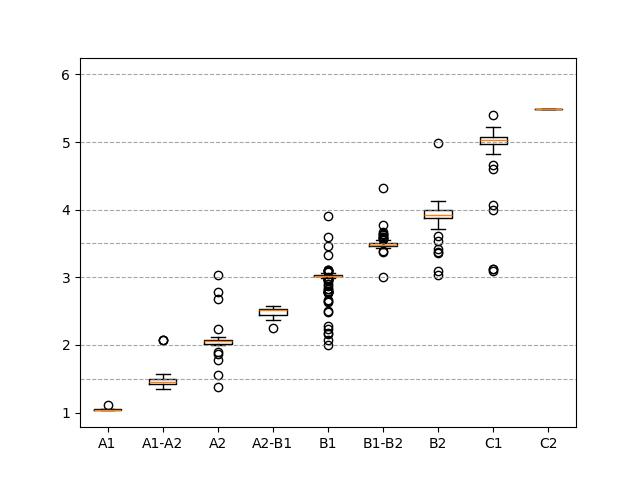}
        \caption{FI train on all data, test on MT,}
        \label{fig:boxplos-RQ1-6}
    \end{subfigure}
    \begin{subfigure}[b]{0.49\textwidth}
    \centering
    \includegraphics[trim=20mm 10mm 20mm \pictop, clip, width=\picwidth\linewidth]{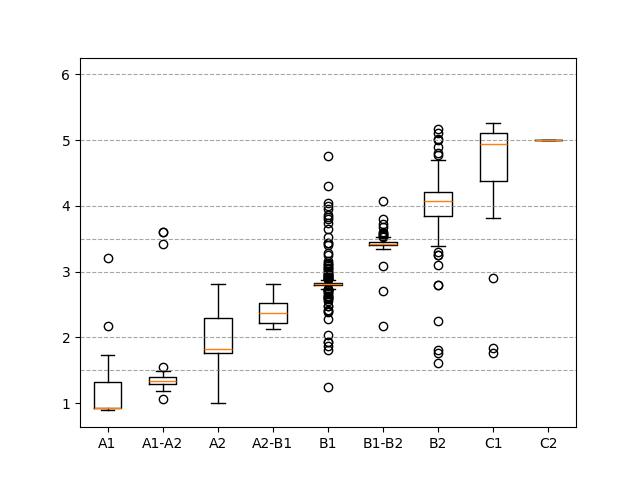}
        \caption{RU train and test on native data}
        \label{fig:boxplos-RQ1-7}
    \end{subfigure}
    \begin{subfigure}[b]{0.49\textwidth}
    \centering
    \includegraphics[trim=15mm 7mm 15mm 18mm, clip, width=\picwidth\linewidth]{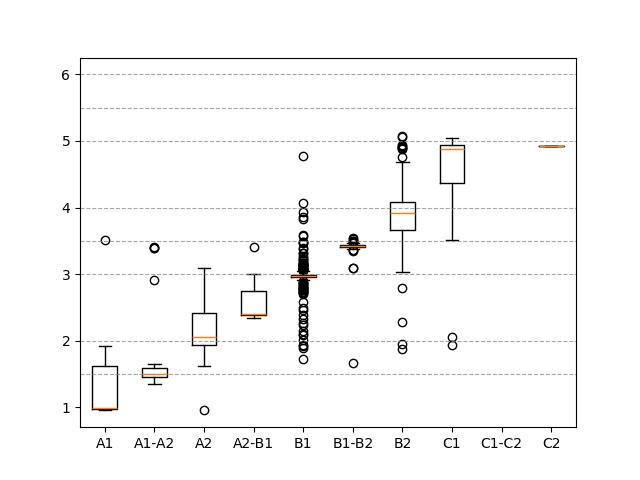}
        \caption{RU train on all data and test on native data}
        \label{fig:boxplos-RQ1-8}
    \end{subfigure}
    \caption{Prediction of difficulty, corresponding to experiments in Table~\ref{tab:results}.}
    \label{fig:boxplos-RQ1}
\end{figure*}

\subsection{Experiments on RQ1}

\begin{table*}[t]
\centering
\resizebox{\textwidth}{!}{
\begin{tabular}{cllllrrrr}
\toprule
\textbf{Exp.} & \textbf{Lang.} & \textbf{Train Set} & \textbf{Dev Set} & \textbf{Test Set} & \textbf{MSE (\%)} & \textbf{RMSE (\%)} & \textbf{MAE (\%)} & \textbf{$R^2$ (\%)} \\
\midrule
\multicolumn{9}{l}{\textbf{Label Injection (whole-label inclusion)}} \\

\midrule
9 & FI & Native + MT (A1--A2) & Native + MT & Native & 29.29 & 54.12 & 41.45 & 84.55 \\
10 & FI & Native + MT (B1--B2) & Native + MT & Native & 6.29 & 25.08 & 11.40 & 96.68 \\
11 & FI & Native + MT (C1--C2) & Native + MT & Native & 16.96 & 41.18 & 25.62 & 91.06 \\
\midrule
\multicolumn{9}{l}{\textbf{Label-wise Proportional Sampling}} \\
\midrule
12 & FI & Native + 20\% MT & Native + MT & Native & 7.83 & 27.97 & 16.81 & 95.87 \\
13 & FI & Native + 40\% MT & Native + MT & Native & 7.20 & 26.84 & 10.66 & 96.20 \\
14 & FI & Native + 60\% MT & Native + MT & Native & 8.22 & 28.68 & 14.72 & 95.66 \\
15 & FI & Native + 80\% MT & Native + MT & Native & 8.71 & 29.51 & 12.74 & 95.41 \\
\BLU{16} & \BLU{FI} & \BLU{Native + 100\% MT} & \BLU{Native + MT} & \BLU{Native} & \BLU{7.57} & \BLU{27.51} & \BLU{9.43} & \BLU{96.01} \\
\bottomrule
\end{tabular}}
\caption{
Model performance on Finnish text difficulty prediction using machine-translated (MT) Russian data.  
\textbf{Label Injection:} Entire CEFR-level subsets (A1--A2, B1--B2, C1--C2) of MT data are injected into the training set.  
\textbf{Label-wise Proportional Sampling:} Russian MT data are sampled proportionally within each CEFR label (20\%–100\%) and added to Finnish native data.  
Note: \BLU{experiment 16} is exactly the same as line 5 in Table \ref{tab:results} (repeated for clarity).
}
\label{tab:combined_label_results}
\end{table*}

Table \ref{tab:results} presents the results in terms of key performance measures---MSE (mean quared error), RMSE (root mean square error), MAE (mean absolute error) and $R^2$ (coefficient of determination)---across different training and testing dataset configurations.  Overall, the results show that incorporating machine-translated data from Russian improves text difficulty prediction for Finnish, compared to training on native data alone, or on translated data alone.

When using only translated Russian data (MT RU→FI), model performance remains modest, with MSE of 24.54\% and $R^2$ of 66.93\% (Figure \ref{fig:boxplos-RQ1}.4). However, combining translated and native Finnish data—while applying dataset balancing—yields a dramatic performance gain , reaching  MSE of 4.22\%, MAE of 8.24\%, and $R^2$ of 94.26\%. The box plots in Figure \ref{fig:boxplos-RQ1}.6 show notably tighter error distributions in this combined setting,\comment{??? please indicate WHICH boxplots: use \ref{subfigure:description} to refer to the SUB-figures } indicating that the model may benefit from both the larger volume and the additional diversity provided by the translated corpus.

The best overall performance is achieved when the model is trained on the native + MT and evaluated on native Finnish data, reaching an $R^2$ of 96.01\% in Figure \ref{fig:boxplos-RQ1}.5.  This demonstrates that machine-translated data not only contributes to improved prediction accuracy in a low-resource setting, but also supports generalization to unseen native Finnish texts.  By contrast, models trained only on translated data perform poorly when tested on native Finnish data---MSE 155.00\%, $R^2$ 18.24\%---which suggests that direct transfer without native exposure is insufficient (Figure \ref{fig:boxplos-RQ1}.3).

Training exclusively on native Finnish data yields strong in-domain performance ($R^2$ 97.30\% in Figure \ref{fig:boxplos-RQ1}.1), confirming the quality of native annotations.  However, the combined model’s comparable accuracy, despite including automatically translated material, shows that translation-based augmentation is an effective strategy for low-resource difficulty prediction.

In sum, the box plots and quantitative results jointly confirm that (1) machine-translated data can substantially enhance low-resource Finnish performance.

\comment{******}

\subsection{Experiments on RQ2}

\comment{??? there is NO REFERENCE to figure 2 \ref{fig:boxplots_RQ2}  -- please add a reference AND references to SUB-figures inside the figure -- HELP THE READER !
}
Table~\ref{tab:combined_label_results} presents the results of ablation studies, combining the native Finnish data with varying amounts of machine-translated (MT) data from the Russian corpus. We explored two strategies: label injection, where entire CEFR-level subsets of translated data (A1–A2, B1–B2, C1–C2) were added to the training set, and label-wise proportional sampling, where translated data were added in increasing proportions (20–100\%) across all levels.

Under the label injection setting, we observe that including B1–B2-level translated texts yields the strongest improvement, achieving $R^2$ of 96.68\% and the lowest overall error rates (Figure \ref{fig:boxplots_RQ2}.2). This suggests that mid-level translated data contribute most effectively to modeling Finnish text difficulty, possibly because they provide balanced lexical and syntactic diversity without overwhelming the model with extreme examples from beginner or advanced levels. In contrast, adding low-level (A1–A2) (Figure \ref{fig:boxplots_RQ2}.1) or high-level (C1–C2)  (Figure \ref{fig:boxplots_RQ2}.3 translated data results in noticeably higher error, indicating limited transferability at the edges of the CEFR scale.

In the experiments with proportional sampling, performance is consistently high across all proportions, with minor fluctuations.  The best result  (Figure \ref{fig:boxplots_RQ2}.5) is obtained at 40\% translated data, reaching $R^2$ of 96.20\%, slightly outperforming full (100\%) augmentation ( (Figure \ref{fig:boxplots_RQ2}.8).  This pattern implies that moderate infusion of translated data effectively regularizes the model, enhancing generalization without introducing domain noise from excessive MT input.

\renewcommand{\thesubfigure}{\arabic{subfigure}}
\def\picwidth{.90}
\def\pictop{23mm} 
\begin{figure*}[htbp]  %
    \centering
    \begin{subfigure}[b]{0.49\textwidth}
    \centering
    \includegraphics[trim=20mm 10mm 20mm \pictop, clip, width=\picwidth\linewidth]{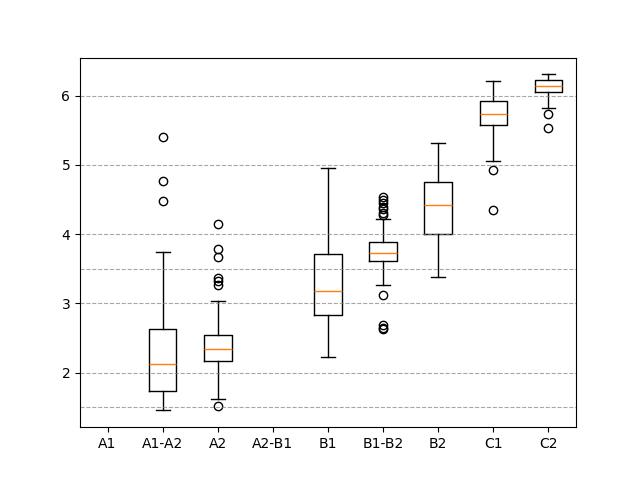}
        \caption{ Exp. 9}
    \end{subfigure}
    \begin{subfigure}[b]{0.49\textwidth}
    \centering
    \includegraphics[trim=20mm 10mm 20mm \pictop, clip, width=\picwidth\linewidth]{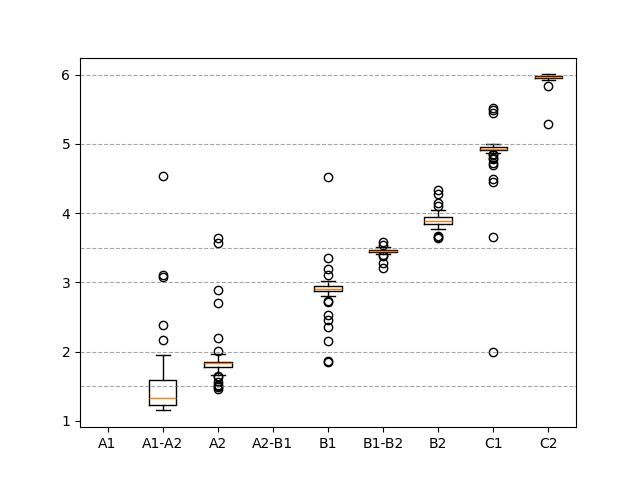}
        \caption{Exp. 10}
    \end{subfigure}
    \begin{subfigure}[b]{0.49\textwidth}
    \centering
    \includegraphics[trim=20mm 10mm 20mm \pictop, clip, width=\picwidth\linewidth]{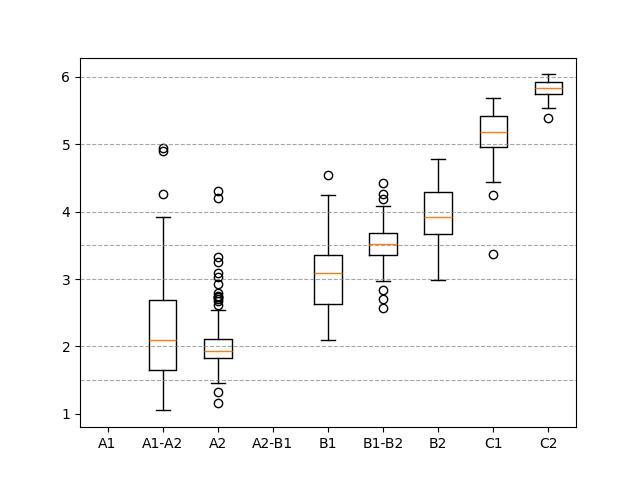}
        \caption{Exp. 11}
    \end{subfigure}
    \begin{subfigure}[b]{0.49\textwidth}
    \centering
    \includegraphics[trim=20mm 10mm 20mm \pictop, clip, width=\picwidth\linewidth]{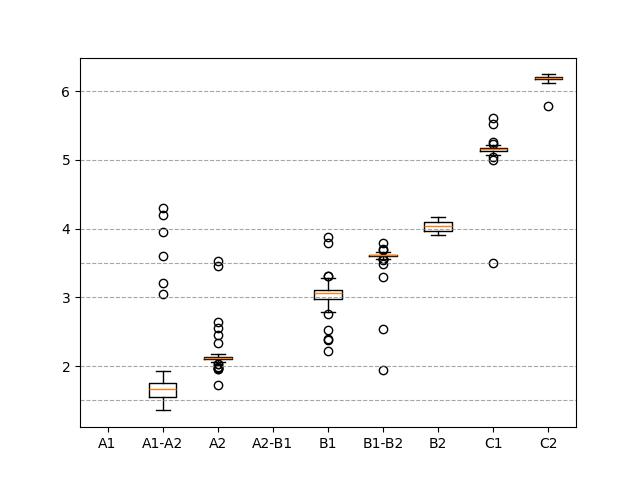}
        \caption{ Exp. 12}
    \end{subfigure}
    
    \vskip\baselineskip
    \begin{subfigure}[b]{0.49\textwidth}
    \centering
    \includegraphics[trim=20mm 10mm 20mm \pictop, clip, width=\picwidth\linewidth]{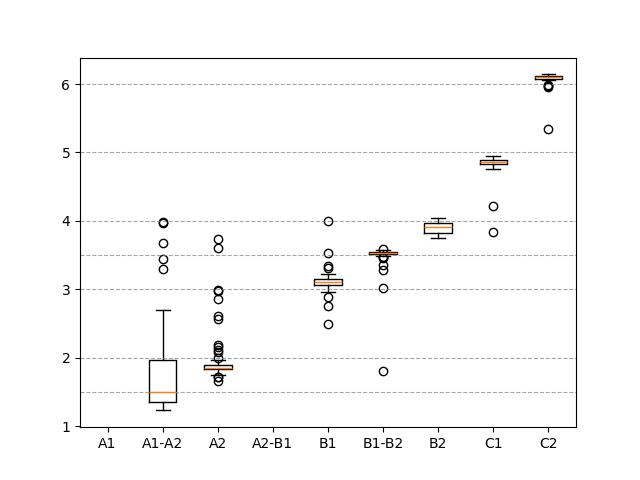}
        \caption{Exp. 13}
    \end{subfigure}
    \begin{subfigure}[b]{0.49\textwidth}
    \centering
    \includegraphics[trim=20mm 10mm 20mm \pictop, clip, width=\picwidth\linewidth]{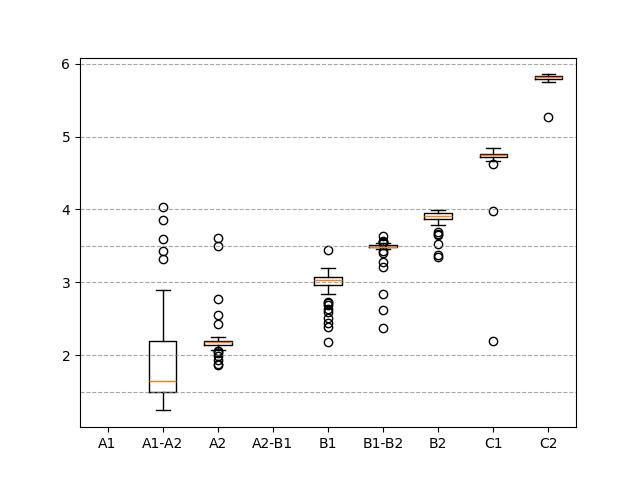}
        \caption{Exp. 14}
    \end{subfigure}
    \begin{subfigure}[b]{0.49\textwidth}
    \centering
    \includegraphics[trim=20mm 10mm 20mm \pictop, clip, width=\picwidth\linewidth]{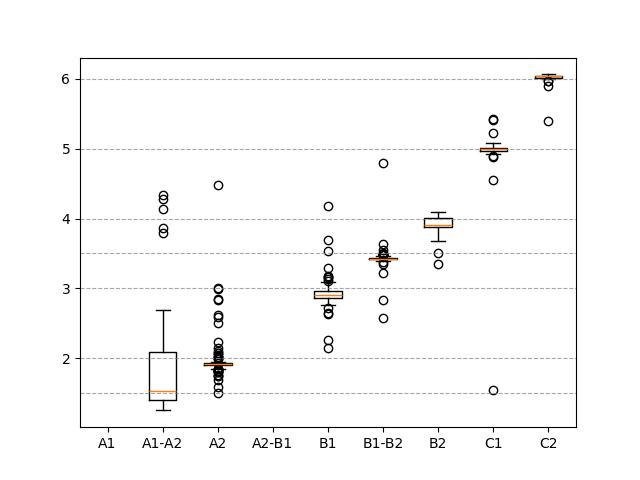}
        \caption{Exp. 15}
    \end{subfigure}
    \begin{subfigure}[b]{0.49\textwidth}
    \centering
    \includegraphics[trim=15mm 7mm 15mm 18mm, clip, width=\picwidth\linewidth]{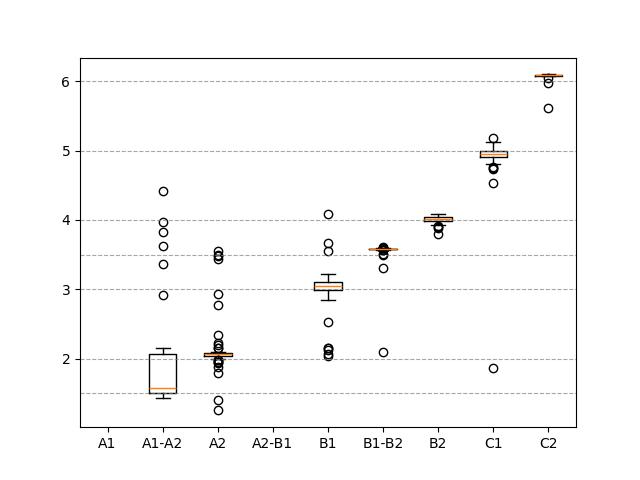}
        \caption{Exp. 16}
    \end{subfigure}
    \caption{Box plots for experiments (Exp. 9–16) showing error distributions across models trained with different proportions and CEFR-level subsets of machine-translated data.}
    \label{fig:boxplots_RQ2}
\end{figure*}

\section{Conclusion}
\label{sec:conclusion}

We set out to explore two research questions:
(1) machine-translated data from a high-resource language can substantially improve prediction of text difficulty in a low-resource setting, and (2) the improvement generalizes across languages, especially when the quantity of translated data are well-balanced across linguistic levels.  

The results confirm that carefully selected MT data can act as a strong proxy for native material in low-resource language modeling.
The experimental findings support both research questions.  Data augmentation via translating text annotated with difficulty from another language can indeed improve the performance of the difficulty model.
A crucial caveat is that we must assure that the particular MT model we use for translation is able to preserve the CEFR level of its input reasonably well.  This is far from a foregone conclusion, since many modern LLMs are trained to ``improve'' on the text while translating, including simplifying the text or making it more ``standard.''  Care must be taken that the MT model preserves the level reasonably accurately, and additional techniques need to be explored to ensure this in a systematic way.

In future work, we plan to pursue several directions.  One plan is to integrate feature-based and Transformer-based models, which could also help with the interpretability of the resulting models in terms of easily understood features.   
We plan to explore whether can provide reasonable guarantees that MT preserves the levels.  This would highlight the importance of our results, since difficulty- and CEFR-annotated data of high quality are very difficult to find, and creating such data is highly resource-intensive.  Data augmentation via MT would allow us to grow our training (and test) datasets considerably, to yield substantial improvements in performance on this complex task.
At the same time will explore multilingual models of difficulty, to study to what extent the transformer can identify language-independent features that impact on text difficulty.

\section{Limitations and Ethical Considerations}
\label{sec:limitations}

While our results show that difficulty models trained on labeled data from multiple languages can be effective, several limitations remain.  First, the models are trained and evaluated on small datasets.  Working only with Finnish may limit generalizability to other languages or domains, and additional languages should be explored.  Second, the mappings that we apply---from continuous regression scores to CEFR levels---introduce discretization errors that may obscure improvements on a more nuanced level.

This work aims at improving language accessibility, particularly for second-language (L2) learners, and seeks to reduce linguistic barriers in education and communication.  However, several ethical considerations must be acknowledged.  First, automated simplification tools may reinforce biases present in the training data, especially if texts from specific groups or dialects are underrepresented.  Second, in general, over-reliance on automated systems may reduce the role of human educators in assessing learner needs, which is not the intent, and is not productive.  Lastly, indiscriminate use or misuse of simplification systems---e.g., to manipulate or oversimplify critical content---can have adverse effects.  We emphasize that these systems should be used as {\em assistive} tools, rather than as replacements for human judgment in the context of education or public communication.


\section*{References}

\bibliographystyle{LREC/lrec2026-natbib}
\bibliography{revita-mtsummit25,custom}

@inproceedings{katinskaia-etal-2025-estimation,
    title = "Estimation of Text Difficulty in the Context of Language Learning",
    author = "Katinskaia, Anisia  and
      Vu, Anh-Duc  and
      Hou, Jue  and
      Vanhatalo, Ulla  and
      Wu, Yiheng  and
      Yangarber, Roman",
    booktitle = "BEA: 20th Workshop on Innovative Use of NLP for Building Educational Applications",
    month = jul,
    year = "2025",
    address = "Vienna, Austria",
    ZZZpublisher = "Association for Computational Linguistics",
    url = "https://aclanthology.org/2025.bea-1.43/",
    doi = "10.18653/v1/2025.bea-1.43",
    XXXpages = "594--611",
    ISBN = "979-8-89176-270-1",

}

@inproceedings{alva-manchego-etal-2025-findings,
    title = "Findings of the {TSAR} 2025 Shared Task on Readability-Controlled Text Simplification",
    author = "Alva-Manchego, Fernando  and
      Stodden, Regina  and
      Imperial, Joseph Marvin  and
      Barayan, Abdullah  and
      North, Kai  and
      Tayyar Madabushi, Harish",
    editor = "Shardlow, Matthew  and
      Alva-Manchego, Fernando  and
      North, Kai  and
      Stodden, Regina  and
      Saggion, Horacio  and
      Khallaf, Nouran  and
      Hayakawa, Akio",
    booktitle = "Proceedings of the Fourth Workshop on Text Simplification, Accessibility and Readability (TSAR 2025)",
    month = nov,
    year = "2025",
    address = "Suzhou, China",
    publisher = "Association for Computational Linguistics",
    url = "https://aclanthology.org/2025.tsar-1.8/",
    doi = "10.18653/v1/2025.tsar-1.8",
    pages = "116--130",
    ISBN = "979-8-89176-176-6",
}

@article{nahatame2026revisiting,
  author  = {Nahatame, Shingo and Yamaguchi, Katsuyoshi},
  title   = {Revisiting Text Readability and Processing Effort in Second Language Reading: {Bayesian} Analysis of Eye-Tracking Data},
  journal = {Language Learning},
  year    = {2026},
  doi     = {10.1111/lang.70011},
  url     = {https://doi.org/10.1111/lang.70011}
}

@article{hurst2024gpt,
  title={{GPT-4o} system card},
  author={Hurst, Aaron and Lerer, Adam and Goucher, Adam P and Perelman, Adam and Ramesh, Aditya and Clark, Aidan and Ostrow, AJ and Welihinda, Akila and Hayes, Alan and Radford, Alec and others},
  journal={arXiv preprint arXiv:2410.21276},
  year={2024}
}

@article{laposhina2020corpus,
  title={A corpus of {Russian} textbook materials for foreign students as an instrument of an educational content analysis},
  author={Laposhina, Antonina},
  journal={Russian Language Abroad},
  volume={6},
  number={283},
  pages={22--28},
  year={2020}
}

@inproceedings{dmitrieva-tiedemann-2021-creating,
    title = "Creating an Aligned {R}ussian Text Simplification Dataset from Language Learner Data",
    author = {Dmitrieva, Anna  and
      Tiedemann, J{\"o}rg},
    editor = "Babych, Bogdan  and
      Kanishcheva, Olga  and
      Nakov, Preslav  and
      Piskorski, Jakub  and
      Pivovarova, Lidia  and
      Starko, Vasyl  and
      Steinberger, Josef  and
      Yangarber, Roman  and
      Marci{\'n}czuk, Micha{\l}  and
      Pollak, Senja  and
      P{\v{r}}ib{\'a}{\v{n}}, Pavel  and
      Robnik-{\v{S}}ikonja, Marko",
    booktitle = "Proceedings of the 8th Workshop on Balto-Slavic Natural Language Processing",
    month = apr,
    year = "2021",
    address = "Kiyv, Ukraine",
    publisher = "Association for Computational Linguistics",
    url = "https://aclanthology.org/2021.bsnlp-1.8/",
    pages = "73--79"
}

@inproceedings{laposhina2018-textometr,
  title={Automated text readability assessment for {Russian} second language learners},
  author={Laposhina, Antonina and Veselovskaya, Tatiana and Lebedeva, Maria and Kupreshchenko, Olga},
  booktitle={Computational Linguistics and Intellectual Technologies},
  pages={403--413},
  year={2018}
}

@techreport{kincaid1975,
  author    = {Kincaid, J. Peter and Fishburne, Robert P. and Rogers, Richard L. and Chissom, Benjamin S.},
  title     = {Derivation of New Readability Formulas (Automated Readability Index, Fog Count and Flesch Reading Ease Formula) for Navy Enlisted Personnel},
  institution = {Naval Air Station Memphis (Research Branch Report 8-75)},
  year      = {1975}
}

@inproceedings{collins-thompson-callan-2004-language,
    title = "A Language Modeling Approach to Predicting Reading Difficulty",
    author = "Collins-Thompson, Kevyn  and
      Callan, James P.",
    booktitle = "Proceedings of the Human Language Technology Conference of the North {A}merican Chapter of the Association for Computational Linguistics: {HLT}-{NAACL} 2004",
    month = may # " 2 - " # may # " 7",
    year = "2004",
    address = "Boston, Massachusetts, USA",
    publisher = "Association for Computational Linguistics",
    url = "https://aclanthology.org/N04-1025/",
    pages = "193--200"
}

@article{martinc-etal-2021-supervised,
    title = "Supervised and Unsupervised Neural Approaches to Text Readability",
    author = "Martinc, Matej  and
      Pollak, Senja  and
      Robnik-{\v{S}}ikonja, Marko",
    journal = "Computational Linguistics",
    volume = "47",
    number = "1",
    month = mar,
    year = "2021",
    address = "Cambridge, MA",
    publisher = "MIT Press",
    url = "https://aclanthology.org/2021.cl-1.6/",
    doi = "10.1162/coli_a_00398",
    pages = "141--179",
}

@inproceedings{gasperin2009,
 author = {Caroline Gasperin and Lucia Specia and Tiago Pereira and Sandra Aluisio},
 title = { Learning When to Simplify Sentences for Natural Text Simplification},
 booktitle = {Anais do VII Encontro Nacional de Inteligência Artificial},
 location = {Bento Gonçalves/RS},
 year = {2009},
 issn = {2763-9061},
 pages = {182--191},
 publisher = {SBC},
 address = {Porto Alegre, RS, Brasil},
 url = {https://sol.sbc.org.br/index.php/eniac/article/view/35695}
}

@inproceedings{aluisio-etal-2010-readability,
    title = "Readability Assessment for Text Simplification",
    author = "Aluisio, Sandra  and
      Specia, Lucia  and
      Gasperin, Caroline  and
      Scarton, Carolina",
    editor = "Tetreault, Joel  and
      Burstein, Jill  and
      Leacock, Claudia",
    booktitle = "Proceedings of the {NAACL} {HLT} 2010 Fifth Workshop on Innovative Use of {NLP} for Building Educational Applications",
    month = jun,
    year = "2010",
    address = "Los Angeles, California",
    publisher = "Association for Computational Linguistics",
    url = "https://aclanthology.org/W10-1001/",
    pages = "1--9"
}

@inproceedings{woodsend-lapata-2011-learning,
    title = "Learning to Simplify Sentences with Quasi-Synchronous Grammar and Integer Programming",
    author = "Woodsend, Kristian  and
      Lapata, Mirella",
    editor = "Barzilay, Regina  and
      Johnson, Mark",
    booktitle = "Proceedings of the 2011 Conference on Empirical Methods in Natural Language Processing",
    month = jul,
    year = "2011",
    address = "Edinburgh, Scotland, UK.",
    publisher = "Association for Computational Linguistics",
    url = "https://aclanthology.org/D11-1038/",
    pages = "409--420"
}

@inproceedings{agrawal-carpuat-2023-controlling,
    title = "Controlling Pre-trained Language Models for Grade-Specific Text Simplification",
    author = "Agrawal, Sweta  and
      Carpuat, Marine",
    editor = "Bouamor, Houda  and
      Pino, Juan  and
      Bali, Kalika",
    booktitle = "Proceedings of the 2023 Conference on Empirical Methods in Natural Language Processing",
    month = dec,
    year = "2023",
    address = "Singapore",
    publisher = "Association for Computational Linguistics",
    url = "https://aclanthology.org/2023.emnlp-main.790/",
    doi = "10.18653/v1/2023.emnlp-main.790",
    pages = "12807--12819",
}

@article{alkaldi-inkpen-2023-readability,
  title   = {Text Simplification to Specific Readability Levels},
  author  = {Alkaldi, Wejdan and Inkpen, Diana},
  journal = {Mathematics},
  volume  = {11},
  number  = {9},
  pages   = {2063},
  year    = {2023},
  doi     = {10.3390/math11092063}
}

@article{sharoff2022neural,
  title={What neural networks know about linguistic complexity},
  author={Sharoff, Serge},
  journal={Russian Journal of Linguistics},
  volume={26},
  number={2},
  pages={371--390},
  year={2022},
}

@article{azpiazu-pera-2019-multiattentive,
    title = "Multiattentive Recurrent Neural Network Architecture for Multilingual Readability Assessment",
    author = "Azpiazu, Ion Madrazo  and
      Pera, Maria Soledad",
    editor = "Lee, Lillian  and
      Johnson, Mark  and
      Roark, Brian  and
      Nenkova, Ani",
    journal = "Transactions of the Association for Computational Linguistics",
    volume = "7",
    year = "2019",
    address = "Cambridge, MA",
    publisher = "MIT Press",
    url = "https://aclanthology.org/Q19-1028/",
    doi = "10.1162/tacl_a_00278",
    pages = "421--436",
}

@inproceedings{vajjala-meurers-2012-improving,
    title = "On Improving the Accuracy of Readability Classification using Insights from Second Language Acquisition",
    author = "Vajjala, Sowmya  and
      Meurers, Detmar",
    editor = "Tetreault, Joel  and
      Burstein, Jill  and
      Leacock, Claudia",
    booktitle = "Proceedings of the Seventh Workshop on Building Educational Applications Using {NLP}",
    month = jun,
    year = "2012",
    address = "Montr{\'e}al, Canada",
    publisher = "Association for Computational Linguistics",
    url = "https://aclanthology.org/W12-2019/",
    pages = "163--173"
}

@techreport{stenner1996,
author="Stenner, A. Jackson",
editor="Fisher Jr., William P.
and Massengill, Paula J.",
title="Measuring Reading Comprehension with the Lexile Framework",
bookTitle="Explanatory Models, Unit Standards, and Personalized Learning in Educational Measurement: Selected Papers by A. Jackson Stenner",
year="2023",
publisher="Springer Nature Singapore",
address="Singapore",
pages="63--88",
isbn="978-981-19-3747-7",
doi="10.1007/978-981-19-3747-7_6",
url="https://doi.org/10.1007/978-981-19-3747-7_6"

}

@inproceedings{dmitrieva-konovalova-2023-creating,
    title = "Creating a parallel {F}innish-{E}asy {F}innish dataset from news articles",
    author = "Dmitrieva, Anna  and
      Konovalova, Aleksandra",
    editor = {Espl{\`a}-Gomis, Miquel  and
      Forcada, Mikel L.  and
      Kuzman, Taja  and
      Ljube{\v{s}}i{\'c}, Nikola  and
      van Noord, Rik  and
      Ram{\'i}rez-S{\'a}nchez, Gema  and
      Tiedemann, J{\"o}rg  and
      Toral, Antonio},
    booktitle = "Proceedings of the 1st Workshop on Open Community-Driven Machine Translation",
    month = jun,
    year = "2023",
    address = "Tampere, Finland",
    XXXpublisher = "European Association for Machine Translation",
    url = "https://aclanthology.org/2023.crowdmt-1.3/",
    pages = "21--26"
}

@phdthesis{Dmitrieva-phdthesis,
author = {Dmitrieva, Anna},
year = {2025},
month = {09},
pages = {},
title = {Resources and Tools for Automatic Text Simplification: Cases of Russian and Finnish}, 
publisher = {University of Helsinki}
}

@inproceedings{imperial-etal-2024-llm,
    title = "{S}pecia{L}ex: A Benchmark for In-Context Specialized Lexicon Learning",
    author = "Imperial, Joseph Marvin  and
      Tayyar Madabushi, Harish",
    editor = "Al-Onaizan, Yaser  and
      Bansal, Mohit  and
      Chen, Yun-Nung",
    booktitle = "Findings of the Association for Computational Linguistics: EMNLP 2024",
    month = nov,
    year = "2024",
    address = "Miami, Florida, USA",
    publisher = "Association for Computational Linguistics",
    url = "https://aclanthology.org/2024.findings-emnlp.52/",
    doi = "10.18653/v1/2024.findings-emnlp.52",
    pages = "930--965",
}

\end{document}